\documentclass[conference]{IEEEtran}
\IEEEoverridecommandlockouts
\usepackage{cite}
\usepackage{amsmath,amssymb,amsfonts}
\usepackage{algorithmic}
\usepackage{booktabs}
\usepackage{algorithm}
\usepackage{graphicx}
\usepackage{textcomp}
\usepackage{xcolor}
\usepackage{amssymb}
\usepackage{amsmath}
\usepackage{makecell}
\newtheorem{theorem}{Theorem}

\usepackage[misc]{ifsym}

\def\BibTeX{{\rm B\kern-.05em{\sc i\kern-.025em b}\kern-.08em
    T\kern-.1667em\lower.7ex\hbox{E}\kern-.125emX}}
\begin{document}

\title{Machine Unlearning Method Based On Projection Residual\\

\author{
    \IEEEauthorblockN{Zihao Cao$^{1}$, Jianzong Wang$^{1(\textrm{\Letter})}$\thanks{$\textrm{\Letter}$ Corresponding author: Jianzong Wang, jzwang@188.com}, Shijing Si$^{1,2}$, Zhangcheng Huang$^{1}$ and Jing Xiao$^{1}$}
    \IEEEauthorblockA{$^1$Ping An Technology (Shenzhen) Co., Ltd., Shenzhen, China}
    \IEEEauthorblockA{$^2$School of Economics and Finance, Shanghai International Studies University, Shanghai, China}
    \IEEEauthorblockA{Email: caozihao1997@gmail.com, jzwang@188.com, shijing.si@outlook.com, \\
    huangzhangcheng624@pingan.com.cn, xiaojing661@pingan.com.cn}
}
}

\maketitle

\begin{abstract}
Machine learning models (mainly neural networks) are used more and more in real life. Users feed their data to the model for training. But these processes are often one-way. Once trained, the model remembers the data. Even when data is removed from the dataset, the effects of these data persist in the model. With more and more laws and regulations around the world protecting data privacy, it becomes even more important to make models forget this data completely through machine unlearning.

This paper adopts the projection residual method based on Newton iteration method. The main purpose is to implement machine unlearning tasks in the context of linear regression models and neural network models. This method mainly uses the iterative weighting method to completely forget the data and its corresponding influence, and its computational cost is linear in the feature dimension of the data. This method can improve the current machine learning method. At the same time, it is independent of the size of the training set. Results were evaluated by feature injection testing (FIT). Experiments show that this method is more thorough in deleting data, which is close to model retraining.
\end{abstract}

\begin{IEEEkeywords}
Machine Unlearning, Projection Residual Method, Text Classification Task, Privacy Protection, Data Security.
\end{IEEEkeywords}

\section{Introduction}
Machine learning should be efficient, requiring less time to run and achieve good accuracy. In addition, machine learning also requires verifiability. That is, it is guaranteed that after data deletion, the machine learning model behaves as if the deleted data was never observed. Such requirements may be mandated by law\cite{kashef2021boosted} (eg the interpretation of the right to forget or erasure in the latest EU law). Sometimes for privacy reasons, such as in some specific projects\cite{sekhari2021remember}\cite{shokri2017membership}, when a user requests to delete their data from their database, the corresponding project needs to delete more than just the data in the managed database. And make sure the data is not learned by any machine learning model. Essentially, under strict auditing, we ensure that the corresponding data has been deleted\cite{guo2019certified}\cite{golatkar2021mixed}, while eliminating the impact of the deleted data on the model. So simply deleting the data is not complete. 

Users need their data to be entirely erased since data security is becoming increasingly crucial. The security of the widely used anomaly detection systems depends on the normal behavior model\cite{tsai2014subunit}\cite{golatkar2020forgetting}\cite{villaronga2018humans} obtained from the training data. An attacker can contaminate the model by contaminating the training data, compromising overall security. For example, Perdisci et al. This indicates that PolyGraph, the worm detection engine\cite{baumhauer2022machine}\cite{shintre2019making}, cannot generate useful worm signatures if the fake network steam is injected with carefully constructed training data. Once contaminated data is identified, the system must completely forget about the data and its pedigree to regain security.

Therefore, machine unlearning has become a hot research topic, the purpose is to allow users to completely delete personal data from the model and training set. Differential privacy and homomorphic encryption used in federated learning\cite{cao2015towards} ensure that people cannot know which data is used by looking at the model, but in fact these privacy-related data\cite{wu2020deltagrad} still contribute to the model. When the contribution degree is 0, the model will not learn anything from the training data. The purpose of machine unlearning is to eliminate the contribution of the target data to the model and completely remove the data from the model and training set.

The above description requires two tradeoffs. Efficiency on the one hand and effectiveness or verifiability on the other: this is because it takes time to optimize the model to reflect the underlying data and to cancel the learning\cite{thudi2022necessity}. Furthermore, there is a trade-off between provability and validity: this is because not learning the deleted data ensures provability, i.e. it can be shown that the model\cite{gupta2020secure} is not using the target data. Deleting the corresponding data means that the model learns from less data, reducing accuracy.

We believe that in order to ensure the privacy, security and usability of the system, the system must be designed so that it can completely and quickly forget sensitive data and its pedigree. We propose a general method for efficient inverse learning without retraining from scratch. The method is a projection residual method based on Newton iteration, which optimizes the model parameters through iteration to make them close to the target model parameters. A model that makes the model completely forget about the target data and its various effects, and is applied to the text classification task of natural language processing.

In summary, the contributions of this paper are:
\begin{itemize}
\item[$\bullet$]In this paper, we mainly use the basis projection residual method to implement the data forgetting task for natural language processing problems. In particular, we use this approach to ensure that the neural network can still perform machine unlearning tasks with low time and space complexity in machine learning, and the effect is similar to model retraining.
\end{itemize}
\begin{itemize}
\item[$\bullet$] Our method adopts the dropout algorithm of the parametric model. The running time of traditional model retraining methods is closely related to the size of the dataset, while the running time of this method is only related to the data dimension. This plays a key role in the crux of modern high-dimensional machine learning.
\end{itemize}

\section{Related work}
\textbf{Model Retraining:} Retraining-based inverse learning methods require retraining after deleting data samples. Most retraining methods focus on particular model classes, employs a warm starting point and a logistic regression approach to data deletion during retraining (Tsai et al., 2014)\cite{deepanjali2021efficient}. The data deletion method proposed by \cite{ginart2019making} in 2019 is mainly used for clustering algorithms. Machine learning has not yet generalized to general model applications. After Papernot et al., the SISA method was proposed. This method divides the training set into shards and shards into slices\cite{schellekens2018quantized}. For each slice, model parameters are recorded after training, dividing each data point into distinct slices and slices. Exclude the corresponding data points during unlearning\cite{tsai2014incremental}\cite{izzo2021approximate}, then retrain the corresponding shards and slices. Essentially, this approach trades space cost for training time cost, however, this approach still does not really reduce the cost of retraining.

\textbf{Summation-based Machine Unlearning:}\cite{koh2017understanding}in 2015 proposed a summation-based machine unlearning method without retraining from scratch\cite{schelter2021hedgecut}. They split the learning algorithm in the system into the form of a number of small sums, where the sum of each part contains all computable transformations of the training data samples. The final learning algorithm relies only on the sum, not the individual data. This data is stored with the trained model. In addition, the entire system\cite{snell2017prototypical} may still require separate data, except for parts of the training model, rather than injecting noise like differential privacy. Then, each time the model is updated by machine learning, the data to be deleted needs to be subtracted from the sum of the parts and then summarized. However, for adaptive models such as neural networks, subtracting one sample from the sum can easily lead to over-learning of irrelevant memories, then significantly reducing utility.

\textbf{Differential Privacy:} \cite{neel2021descent} in 2019 proposed a new concept of machine unlearning influenced by differential privacy methods, treating data deletion as an online problem. They used a simple but long-standing k-means clustering problem\cite{bertram2019five} to de-emphasize effective learning case studies. Two efficient deletion algorithms are proposed, Q-k-means\cite{aldaghri2021coded} and DC-k-means\cite{ghorbani2019data}. These methods (in some cases) achieve the best delete efficiency. Meanwhile they impose requirements on the output distribution of the machine learning algorithm, which should be similar to the output distribution obtained by the target learning algorithm. The dataset used by the learning algorithm is and no target points are removed.

The method adopted in this paper avoids the problem of excessive resource consumption caused by retraining. We adopted the projected residual method\cite{chaudhuri2011differentially}\cite{chen2021machine} based on Newton's method. This method only processes data that needs to be deleted, and does not over-learn irrelevant memories. The time complexity of traditional methods is related to the size of the deleted data, but this method is only dimension dependent and does not grow exponentially\cite{schelter2019amnesia}. In terms of removal accuracy, this method is optimal for some gradient-based updates.

\section{Methodology}
\label{sec:format}

\subsection{Parameter Description}

In this part, we explain what the parameters needed for this procedure mean:

\begin{itemize}
\item[$\bullet$] $D^{full} = {(x_i,y_i)}^n_{i=1}$ is the full training dataset. The feature vector $x$ in the dataset, which we regard as irrelevant. That is, any set not exceeding $D^{full}$ can be considered linearly independent.
\end{itemize}
\begin{itemize}
\item[$\bullet$] $X = [x_1,x_2,...,x_n]^T \in R^{n \times d}$ is the specific data information of $D^{full}$, expressed in the form of matrix, where the $x_i^T$ are the corresponding eigenvectors.
\end{itemize}
\begin{itemize}
\item[$\bullet$] $Y = (y_1,y_2,...,y_n)^T \in R^n$ is the label for $D^{full}$
\end{itemize}
\begin{itemize}
\item[$\bullet$] $\theta \in R$ denote the model parameters
\end{itemize}
\begin{itemize}
\item[$\bullet$] $\theta^{full} = argmax_{\theta}L^{full}(\theta)$ is the model parameter when fitting to the full data set. We refer to models that use these parameters as complete models. 
\end{itemize}
\begin{itemize}
\item[$\bullet$] $L^{full}(\theta) = \sum_{i=1}^n \iota(x_i.y_i;\theta) +\frac{\lambda}{2}\left \| \theta \right\|^2_2$ is the loss function of full datasets. The single point loss function is quadratic loss for linear regression $(\frac{1}{2}(\theta^Tx_i - y_i)^2)$. 
\end{itemize}
\begin{itemize}
\item[$\bullet$] $D^k = {(x_i,y_i)}^n_{i=k+1}$ is the processed dataset, in which $k$ feature points are deleted. Suppose these deleted points are the top $k$.
\end{itemize}
\begin{itemize}
\item[$\bullet$] $L^{k}(\theta) = \sum_{i=k+1}^n \iota(x_i.y_i;\theta) +\frac{\lambda}{2}\left \| \theta \right\|^2_2$ is the loss function of the dataset that has deleted $k$ points. We require that the regularization strength is independent of sample size.
\end{itemize}
\begin{itemize}
\item[$\bullet$] $\theta^{k} = argmax_{\theta}L^{k}(\theta)$ are the model paramteres when fitted to the dataset that has deleted $k$ points. The regularization strength of the loss function is independent of the sample size.
\end{itemize}
\begin{itemize}
\item[$\bullet$] $y^k_i = (\theta^{k})^{T}x_i$ is the prediction label at the ith point after deleting data.

\end{itemize}

\subsection{Machine Unlearning Based On Newton} Anti-learning method based on summation is a kind of implementation based on statistical query learning. \(G_i = \sum_{x \in D} g_i(x)\)  represents the sum of the transformations of the training samples in the \(i_th\) iteration, Where \(g_i(x)\) is the transformation of a specific sample $x$, training set is D. One model trained through $k$ iterations can be expressed as: \(f_k = Learnt(G_1,G_2,...,G_k)\). When sample \(D'\) needs to be unlearned, the calculation equation of summation-based MU(machine unlearning) is:
\begin{equation}
f'_k = Learnt(G_1-G_1',G_2-G_2',...,G_K-G_K') \label{ZZ1}
\end{equation}
\begin{equation}
G_I' = \sum_{x \in D'} g_i(x) \label{ZZ2}
\end{equation}

\(f_k\) is the updated model with the \(D'\) memory removed. This approach works very well under non-adaptive learning models. However, in the adaptive learning neural network, small parameter changes will lead to deviations in the subsequent training results, thus affecting the final effect.

Our dataset using the Newton iteration-based projected residual method consists of existing data and synthetic data. The intuition is the following: if we don't know \(\theta_k\), then at \(i = 1,... , k\) we should expect our parameters to be closer to \(\theta_k\). Here \(\theta_k\) is the new parameter after the model is retrained, and it is also the target parameter that our method expects to approximate. Because \(\theta_k\) will use gradient descent method at point \((x_i, y^k_i)\) to achieve the minimum loss.

The computational cost of the experiment in this paper is mainly divided into two parts, one is to update the parameters after deleting the request, and the other is to complete the rapid redeployment of the deleted model. When discussing the computational cost, we only consider the computational cost of each method at the time of removal request, and do not calculate the cost of precomputing\cite{gupta2021adaptive} a reasonable amount of partial, (i.e. without knowing the $k$ points to delete calculations completed below). For the parts that are not considered, we just exclude trivial but too expensive methods. 

The experiments in this paper mainly focus on completing one batch delete request. Although the number of requests is small, our method is different and more critical than traditional methods like retraining. Extending this approach to the current fully online working environment, multiple batch deletion requests may be received in a task, and the pre-training required between each request will also play a key role. We will discuss the machine learning effects of linear regression and bidirectional LSTM models\cite{pruthi2020estimating} in the main body of this article, looking at the completion under one batch request.

\textbf{Newton Iteration Method:}  In recent years, some work has tried to use the Newton iteration method to obtain the objective function based on the loss function $L$ of the deleted data set $D$, using the quadratic function of $L$ to move to the nearest minimum\cite{graves2021amnesiac}.

\begin{equation}
\theta_{new} = \theta_{full} - \nabla^2_{\theta}L^k(\theta_{full})^{-1}\nabla L^k(\theta_{full})  \label{ZZ3}
\end{equation}

In some cases, the results of Newton's iteration method are approximately exact. For the model parameter \(\theta\), the loss function \(L^k\) can be considered to be approximately the same when it is quadratic. Therefore, Newton's method is a simplified, repeatable model retraining method not only in linear regression problems, but also in neural networks. And this method has higher accuracy.

We map it to the NLP text classification task according to the projected residual principle of projected residuals. This method is mainly used for linear models, when retraining deep neural networks, all but the last layer are treated as fixed feature maps\cite{shoeybi2019megatron}. Only the last layer is retrained, which also serves the purpose of neural network machine learning. In\cite{ginart2019making}, the authors retrained their model to calculate Shapley values for the data and to measure the contribution of each data point to the overall accuracy of the model. They have demonstrated that we can obtain precise results in this situation by training the model's last fully linked layer. As a result, the LSTM model may also be used with this approach.

The projected residuals used in this experiment can be derived as logistic regression. It makes use of logistic models, which iteratively weighted least squares may be used to train. In actuality, the weighted least squares problem solution constitutes the Newton step\cite{cao2018efficient} of logistic regression. But this paper focuses on linear regression and neural networks implemented by the Newton approximation. Logistic regression is not something to discuss specifically.

\subsection{Influence Method}
Estimating the impact of various training points on model predictions has been the subject of recent investigations. The strategy was developed despite being used for several objectives. For instance, approximative data removal may be done by modifying cross-validation and interpretation\cite{wager2018estimation}. We may consider the model parameters \(\theta\) as a function of the data weights if we make the right assumptions about the loss function: \(\theta(w)=argmin_{\theta}\sum_{i=1}^{n}w_il(x_i,y_i;\theta)\) .The influence function approach uses linear approximations to \(\theta(w)\) to predict \(\theta^k\). The linear approximation is given by:

\begin{equation}
\theta^{inf}=\theta^{full}-[\nabla^2_{\theta}L^{full}(\theta^{full})]^{-1}\nabla_{\theta}L^k(\theta^{full})
\label{ZZ4}
\end{equation}

\subsection{Algorithm Flow Derivation}
The specific algorithm implemented in this article is as follows: Assuming that we can calculate \(y^k_i\) efficiently without knowing \(\theta_k\), then the gradient loss of the composite point \((x_i,y^k_i)\) is \(\nabla_\theta L^{(x_i,y^k_i)} = \sum_{i=1}^{n}(\theta^Tx_i - y^k_i)^2x_i\). Then replace \(y^k_i\) with \(\theta^{kT}x_i\) and set \(\theta = \theta_{full}\), where \(\theta_{full}\) is the result obtained by training the model on the complete dataset. The new loss function is

\begin{equation}
\nabla_\theta L^{(x_i,y^k_i)} = (\sum_{i=1}^{n}x_ix^T_i)(\theta_{full} - \theta_k) \label{ZZ5}
\end{equation}

$\sum_{i=1}^{n}x_ix^T_i$, we can compute pseudoinverses quickly using this technique. The exact steps involved in the projected residual update are described in Algorithm 1. The aforementioned approach only has an  \(O(k^2d)\) computation time complexity, where $d$ is the number of dimensions in the data. It is obvious that the linear combination of deleted \(x_i\)) has been improved the most by the projection residual update compared to the other parameter updates.

\begin{algorithm}[htb]
\caption{Projected residual update}
\label{alg:Framwork}
\begin{algorithmic}[1] %这个1 表示每一行都显示数字
\REQUIRE ~~\\ %算法的输入参数：Input
    The data matrix for full datasets(its rows are the feature vectors), $X$;\\
    The response vector for full datasets, $Y$;\\
    The hat matrix, $H$;\\
    The model parameter when fitting to the full data set, $\theta^{full}$\\
    Number of deleted data, $k$
\ENSURE ~~\\ %算法的输出：Output
    Ensemble of classifiers on the current batch, $E_n$;
    \STATE $y_1,y_2,...,y_k \leftarrow$ DK$(X,Y,H,k)$;
    \STATE $S^{-1} \leftarrow pseudoinverse (\sum_{i=1}^{n}x_ix^T_i)$ ;
    \STATE $\nabla L \leftarrow \sum_{i=1}^{k}(\theta^{full T}x_i - y_i)x_i$; 
\RETURN $\theta^{full} - FASTMULT(S^{-1},\nabla L)$ %算法的返回值
\end{algorithmic}
\end{algorithm}

The calculation process of the pseudo-inverse in Algorithm 1 is described in the following Algorithm 2:

\begin{algorithm}[htb]
\caption{Compute the pseudoinverse of $\sum_{i=1}^{k}x_ix_i^{T}$}
\label{alg:Framwork}
\begin{algorithmic}[1] %这个1 表示每一行都显示数字
\REQUIRE ~~\\ %算法的输入参数：Input
    The whole datasets' data matrix, $x_1,x_2,...,x_k$
\ENSURE ~~\\ %算法的输出：Output
    $\lambda_1^{-1},v_1,...,\lambda_k^{-1},v_k$
\STATE $c_1,...,c_k,u_1,...,u_k \leftarrow $Gram-Schmidt$(x_1,...,x_k):x_i = \sum_{j=1}^{k}c_{ij}u_j $
\STATE $C \leftarrow \sum_{i=1}^{k}c_ic_i^{T}$
\STATE $\lambda_1,a_1,...,\lambda_k,a_k \leftarrow $Eigendecompose$(C)$
\FOR {$i=1,2,...,k$ }
\STATE $v_i \leftarrow \sum_{j=1}^{k}a_{ij}u_j$ 
\ENDFOR %算法的返回值
\RETURN  $\lambda_1^{-1},v_1,...,\lambda_k^{-1},v_k$ %算法的返回值
\end{algorithmic}
\end{algorithm}

For any dataset $D = (x_i,y_i)$, its loss function can be defined as \(L^D(\theta) = \sum_{i=1}^{n}\frac{1}{2}(\theta^Tx_i - y^k_i)^2x_i\) Through the formula shown below, the gradient update \(\theta_{update}\) of the model parameter \(\theta^{full}\) of the complete data set is defined as. Theorem 1, our major theorem, describes the outcomes of applying the residual update.

\begin{equation}
\theta_{i+1} = \theta_i + \alpha_i\nabla_{\theta}L^{D_i}(\theta_i) \label{ZZ6}
\end{equation}

After removing $k$ data, we take gradient steps in the specified direction of the remaining data. That is, we update

\begin{equation}
\theta^{res}=\theta^{full}-\alpha\sum_{i=1}^{k}(\theta^{full T}x_i - y_i)x_i
\label{ZZ7}
\end{equation}

\begin{theorem}
The computational cost of Algorithm 1 calculation $\theta^{res} = \theta^{full} + proj_{span(x_1,...,x_k)}(\theta^k - \theta^{full})$ is $O(k^2d)$
\end{theorem}

Typically, the step size is specified by the vector parameter $\alpha$. For our needs, we'll swap out for a "step matrix" $A$. When we replace the learning rate $\alpha$ in (7), the equation is rewritten as:

\begin{equation}
\begin{aligned}
\theta^{res}&=\theta^{full}-A\sum_{i=1}^{k}(\theta^{full T}x_i - y_i)x_i \\
&=\theta^{full}-A\sum_{i=1}^{k}(\theta^{full T}x_i - \theta^{k T}x_i)x_i \\ 
&= \theta^{full}-A\sum_{i=1}^{k}x_ix_i^{T}(\theta^{full} - \theta^k)
\end{aligned}
\label{ZZ8}
\end{equation}

Then we let $N = \sum_{i=1}^{k}x_ix_i^{T}$, also note that the $range(N) = span(x_1,...,x_k) \triangleq V_k$, With the form of N, We may quickly determine its eigendecomposition by computing $N = E \Lambda E^T$,  $\Lambda = diag(\lambda_1,...,\lambda_k,0,...,0)$, $\lambda_1,...,\lambda_k$, are the nonnegative eigenvector of $N$. We then identify:

\begin{equation}
A = E\Lambda^{'}E^T
\end{equation}

This selection of $A$ results in $AN = \sum_{i=1}^{k}v_iv_i^{T} = proj_{v_k}$,and therefore the update (8) is equivalent to

\begin{equation}
\theta^{full} + proj_{v_k}(\theta^k - \theta^{full})
\end{equation}

In Algorithm 1, we use the existing \(pseudoinv\) method for calculation, which involves gram calculation. Gram-i-th Schmidt's step gives us the following results:
\begin{equation}
x_i = (x_i^T u_1)u_1 + ... + (x_i^T u_{i-1})u_{i-1} + ||w_i u_i||  \label{ZZ9}
\end{equation}

\begin{equation}
c_{ij}=\left\{
\begin{aligned}
x_i^Tu_j & & {1 \leq j < i} \\
||w_i|| & & i = j \\
0 & & j > i
\end{aligned}
\right.
\end{equation}

The asymptotic time complexity of this phase can be reduced by storing these coefficients as they are computed in the Gram-Schmidt process.

\begin{algorithm}[htb] 
\caption{Quick division with pseudoinverse} 
\label{alg:Framwork} 
\begin{algorithmic}[1] %这个1 表示每一行都显示数字
\REQUIRE ~~\\ %算法的输入参数：Input
$S^{-1},\bigtriangledown L$  when the low-rank version of $S^{-1}$ must be provided $S^{-1} = \sum^k_{i=1}\lambda_i^{-1}v_iv_i^T$
\ENSURE ~~\\ %算法的输出：Output
\RETURN $\sum^k_{i=1}(\lambda_i^{-1}(v_i^T \bigtriangledown L))v_i$
\end{algorithmic}
\end{algorithm}

The equivalent set is made up of the set of removed points from the original dataset. The set's feature vector is called \(S = {xi}^{k}_{i = 1}\). If any gradient-based update of \(\theta^{full}\) is what \(\theta^{update}\) is, we may determine:
\begin{equation}
   ||\theta^{k}-\theta^{res}|| < ||\theta^{k} - \theta^{update}||  \label{ZZ10}
\end{equation}

Given that any point $(x_i, y)$'s gradient of squared loss is a scalar multiple of $x_i$, update $\theta^{full}  -  \theta^{update} \in span(x1_, ... ,x_k)$. When we use the two-way LSTM model to process text classification tasks, The algorithm flow can be obtained in Algorithm 3.

\begin{algorithm}[htb]
\caption{DK(delete k points) prediction}
\label{alg:Framwork}
\begin{algorithmic}[1] %这个1 表示每一行都显示数字
\REQUIRE ~~\\ %算法的输入参数：Input
    The data matrix for full datasets(its rows are the feature vectors), $X$;\\
    The response vector for full datasets, $Y$;\\
    The hat matrix, $H$;\\
    The model parameter when fitting to the full data set, $\theta^{full}$\\
    Number of deleted data, $k$
\ENSURE ~~\\ %算法的输出：Output
    After removing the k-point dataset, the model's prediction $Y^k$;
    \STATE $R \leftarrow Y_{i:k} - X_{i:k}\theta^{full}$;
    \STATE $D \leftarrow diag(((1-H_ii)^{-1})_{i=1}^k)$; 
\label{code:fram:add}
    \STATE $T_ij \leftarrow 1(i\ne j)\frac{H_ij}{1-H_jj}$; 
    \STATE $T \leftarrow (T_ij)^k_{i,j=1}$; 
    \STATE $Y_k \leftarrow Y_i(I-T)^{-1}DR$; 
\RETURN $Y^k$ %算法的返回值
\end{algorithmic}
\end{algorithm}

\begin{algorithm}[htb]
\caption{Projected residual update of neural networks}
\label{alg:Framwork}
\begin{algorithmic}[1] %这个1 表示每一行都显示数字
\REQUIRE ~~\\ %算法的输入参数：Input
    The data matrix for full datasets(its rows are the feature vectors), $X$;\\
    The response vector for full datasets, $Y$;\\
    The hat matrix, $H$;\\
    The model parameter when fitting to the full data set, $\theta^{full}$\\
    Number of deleted data, $k$
\ENSURE ~~\\ %算法的输出：Output
    Ensemble of classifiers on the current batch, $E_n$;
\FOR {each round $t=1,2 \dots$ }
\STATE $w_i \leftarrow embedding,Bidirectional$ 
\STATE $H^{-1} \leftarrow \sum_{i=1}^{n}x_ix^T_i$ 

\STATE $\nabla L^{D_i}(w_i) \leftarrow \sum_{i=1}^{n}(w^Tx_i - y^k_i)^2x_i $
\STATE $w_{i+1} = w_i - \alpha\nabla L^{D_i}(w_i)$
\STATE \textbf{return} $\nabla w$
\ENDFOR %算法的返回值
\end{algorithmic}
\end{algorithm}

Our method depends on the accuracy of the $y_i^k$ computation. By creating the computation of A that permits the computing of a leave-one-out residual $y$, Algorithm 4 derives a well-known statistician's finding. According to Algorithm 4, $X_{i:k}$ stands for the first $k$ rows of $X$, $Y_{1:k}$ for the first $k$ entries of $Y$, and $H$ for the data's hat matrix. It is possible to omit these steps from the computation of the total processing cost of Algorithm 4 since the residual $r_i = y_i - x_i^T\theta^{full}$ and hat matrix $H$ may be calculated before the request is eliminated.

We add this method to the Bidirectional LSTM method for text classification for machine unlearning training, iterating weights over the last layer of network Bidirectional to eliminate the effect of deleted data. Algorithm 5 describes the specific process. We adopted the feature Injection test (FIT) method to evaluate the effect of the model.

Additionally, we discovered that this iterative method does not update the parameters when we attempt to eliminate points with a high norm in a text classification test. Additionally, the update gets smaller the bigger the features that were eliminated are. An accurate estimate of the Hessian for the leave-one-out loss is a Hessian whose impact on method performance relies on the entire loss\cite{wang2019neural}. If we already have access to the inverse of the Hessian, then the Hessian-gradient product is the method's bottleneck. However, when the size of the outliers grows, a clearly defined finite limit is approached by the precise parameter $\theta_{full} - \theta_1$ update vector.

\section{experiment}
\subsection{Experiment Setup}
\textbf{Dataset:} In the file classification of the news article example, there is this many-to-one relationship. The input is a sequence of words, and the output is a single class or label. This experiment is to complete the classification of sample articles. The data set we used is public data set on BBC news articles\footnote{https://www.kaggle.com/yufengdev/bbc-te}. The total number of labels in the data is 2226. In addition, we used synonyms to replace a single word, delete a word, change the word order, etc. to expand the text data, and build a synthetic text data set for training. The total number of labels in synonyms dataset is 2326

\textbf{Evaluation Metrics:} We employed the Feature Injection Test (FIT) approach published by Zachary Izzo et al. The customer requests the model to forget any learned local dependencies when they need their data removed from the model, which justifies this evaluation technique. There is a clear signal in our dataset that we need the model to pick up on\cite{mahadevan2022certifiable}. In particular, we updated the data with a new feature. All of the data points, with the exception of a very small number, are zero and entirely connected with the label we aim to forecast. We demand that the model learn a weight with an absolute value that is significantly higher than zero for a linear classifier. The removal of this specific subgroup, however, necessitates some stringent regularization. A weight for a linear classifier with an absolute value that is considerably higher than zero is what we require the model to learn. However, after eliminating this particular subgroup, any stringent regularization results in the weight of this feature being set to 0 in the whole retrained model\cite{yeom2018privacy}. In essence, the additional features we added are removed in line with the removal of the target data. In the process of retraining, this added feature is decreased to zero.

\subsection{Experiment Based On Linear Regression}
The dataset for linear regression was manually created and consists primarily of customer reviews of a product. Since the data is well-structured, linear classification benefits from it. We decided to take advantage of the dataset's user reviews. From just one review sample, we created a vocabulary of 1600 frequent terms in the dataset. Then, for each review in the dataset, we encode the number of times each word in the lexicon appears in that review as a vector of counts. Reviews four and five are deemed favorable; the rest are unfavorable\cite{baldi2014searching}. We set the score cutoff to zero to transform the regression model's predictions into a binary classifier. The forecasts of the positive class are those that are more than zero\cite{bourtoule2021machine}, while the predictions of the negative class are those that are less than zero.

In this experiment, where there were 100 such experiments, the total running time of retraining included the average running time of each approach. In the following description, we will abbreviate the influence method as inf. Standard errors were never within significant digits of the mean in any of the examples (all standard errors were on the order of \(10^{-4}\) or less). The related findings are shown in Table 1, where data is randomly destroyed to varying degrees.

\begin{table}[!htbp]
    \centering
    \caption{FIT assessment results on linear regression$(10^{-4})$}
    \begin{tabular}{ccccc}
        \toprule
        Method & d = 1000 & d = 1500 & d = 2000  & d = 3000\\
        \midrule
        k = 1(INF) & 79 & 50 & 42  &29\\
        \textbf{k = 1(Ours)} & \textbf{60} & \textbf{15} & \textbf{9}  &\textbf{4}\\
        k = 5(INF) & 99 & 60 & 47  &30\\
        \textbf{k = 5(Ours)} & \textbf{125} & \textbf{43} & \textbf{25}  & \textbf{12} \\ 
        k = 10(INF) & 103 & 64 & 51  &31 \\
        \textbf{k = 10(Ours)} & \textbf{155} & \textbf{43} & \textbf{27}  & \textbf{12} \\ 
        k = 30(INF) & 115 & 70 & 58  &35 \\
        \textbf{k = 30(Ours)} & \textbf{405} & \textbf{141} & \textbf{79} & \textbf{32} \\ 
        k = 50(INF) & 132 & 75 & 63  &40 \\
        \textbf{k = 50(Ours)} & \textbf{805} & \textbf{294} & \textbf{179}  & \textbf{38} \\ 
        k = 100(INF) & 152 & 89 & 77 &47 \\
        \textbf{k = 100(Ours)} & \textbf{1511} & \textbf{593} & \textbf{296}  & \textbf{61} \\ 
        k = 200(INF) & 316 & 172 & 149  &77 \\
        \textbf{k = 200(Ours)} & \textbf{3492} & \textbf{1276} & \textbf{606}  & \textbf{102} \\
        \bottomrule
    \end{tabular}\label{tb:Fitness_formations1}
\end{table}

As we can see from the table, the running time of the method adopted in this experiment is particularly favorable for high-dimensional data and relatively small $k$.

\subsection{Experiment Based On Neural Networks}
In the experimental part based on the Newton iteration method, the data set is news summaries in the Curation Corpus data set, and some news summaries data are also randomly deleted in this part. We adopt a 6-layer LSTM model to segment text topics by summarizing. Extreme situations (deletion group 1) and small-scale data deletion are both tasks. The experimental results show that our method is much faster than model retraining.

Table 2: When the data dimension is 1000, the corresponding time-consuming of different methods is more. After the corresponding projection residual method is improved, it can be seen that the time is significantly reduced. The accuracy of the original LSTM model training is $93.48\%$. In contrast, the result accuracy of the projected residual method is close to $93\%$, which does not affect the model effect. 

\begin{table}[!htbp]
    \centering
    \caption{Time Consumption}
    \begin{tabular}{cccc}
        \toprule
        methods & Min Time 	& Max Time	 & total\\
        \midrule
        Retrain 	& 160s	& 200s	     & 1720s  \\
        Retrain(Synthetic data) 	& 160s	& 200s	     & 1780s  \\ 
        SISA & 145s	& 180s	     & 1680s  \\ 
        INF 	& 130s	& 170s	     & 1550s  \\ 
        \textbf{Projection residual(Ours)}  &  \textbf{110s}		& 	\textbf{150s}	 &   \textbf{1350s} \\
        \bottomrule
    \end{tabular}\label{tb:Fitness_formations2}
\end{table}

After that, we analyze the machine unlearning results using the FIT approach. Since none of the standard errors among them were $10^{-4}$ or greater and none were within the effective digits of the mean, we decided to exclude the error from the result to make it more comprehensible.

Table 3: Comparison of the effects of the Newton iteration-based projected residual method and the INF method. \(d\) in the table is the dimension size of the data. When the deletion amount is small, the FIT value is also close to 0. However, with the gradual increase of the deletion amount, the FIT value also increased, and the increase rate was also obvious.

\renewcommand{\arraystretch}{0.8}
\begin{table}[!htbp]
    \centering
    \caption{FIT assessment results$(10^{-4})$}
    \begin{tabular}{cccccc}
        \toprule
        Number of deletions & 1 	& 5	 &  10  & 20  & 50\\
        \midrule
        INF & 85 & 98 & 187 & 350 & 921\\
        \makecell{Projection residual\\(d = 1000)}& \textbf{81} & \textbf{90} & \textbf{157} & \textbf{325} & \textbf{807}\\
        INF & 40 & 45 & 76 & 174 & 381\\
        \makecell{Projection residual\\(d = 1500)}& \textbf{34} & \textbf{39} & \textbf{46} & \textbf{102} & \textbf{301}\\
        INF & 17 & 24 & 43 & 97 & 281\\
        \makecell{Projection residual\\(d = 2000)}& \textbf{14} & \textbf{21} & \textbf{25} & \textbf{59} & \textbf{176}\\ 
        INF & 8 & 18 & 20 & 44 & 82\\
        \makecell{Projection residual\\(d = 2500)}& \textbf{5} & \textbf{12} & \textbf{14} & \textbf{31} & \textbf{68}\\ 
        \bottomrule
    \end{tabular}\label{tb:Fitness_formations3}
\end{table}

In addition, we can see from the table that when the initial data dimension is 1000, the FIT value of deleting a piece of data is 0.0081. When the data dimension is 1500, the FIT value drops sharply to 0.024. In addition, we compare FIT, SISA and the method in this paper in different data dimensions, and the results are shown in Figure 1. Since SISA is a method based on retraining, it can be seen that our method is an approximate and retraining method in effect, but the efficiency is significantly improved.

\begin{figure}[h]
\center{\includegraphics[width=8cm] {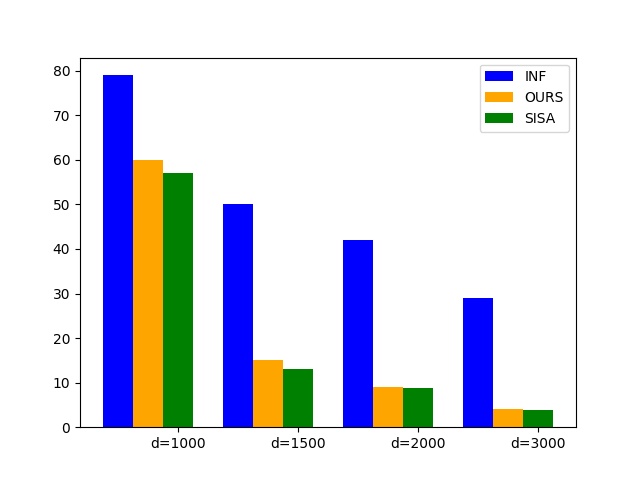}} 
\caption{\label{1} FIT assessment results} 
\end{figure}

Table 4: Average results based on combined data FIT (50 trials). where special weights are the scores obtained as baseline weights (lower is better). For different group sizes $(k)$ and sparse value distributions $(p)$, the outcome is $d = 1500$. These outcomes, as we can see, support our hypothesis that the experimental strategy works best in the sparse state.

\begin{table}[!htbp]
    \centering
    \caption{Mean results for the FIT}
    \begin{tabular}{cccc}
        \toprule
        k value & p = 0.5 	& p = 0.25 &  p = 0.1 \\
        \midrule
        k = 10(INF)	& 1.19	& 1.09	     & 1.10  \\
        \textbf{k = 10(Ours)} & \textbf{1.12} & \textbf{0.98}	   & \textbf{1.07}  \\
        k = 50(INF) 	& 0.96	& 1.05	     & 3.11  \\ 
        \textbf{k = 50(Ours)} 	& \textbf{0.96}	& \textbf{0.98}	& \textbf{0.65}  \\
        k = 100(INF)  &  0.88		& 	1.02	 &   1.22 \\
        \textbf{k = 100(Ours)}  &  \textbf{0.79}	& 	\textbf{0.62}	 &   \textbf{0.33} \\
        \bottomrule
    \end{tabular}\label{tb:Fitness_formations4}
\end{table}

As data dimensions continue to grow, FIT will continue to decline. The changes are shown in Figure 2. 

\begin{figure}[h]
\center{\includegraphics[width=8cm] {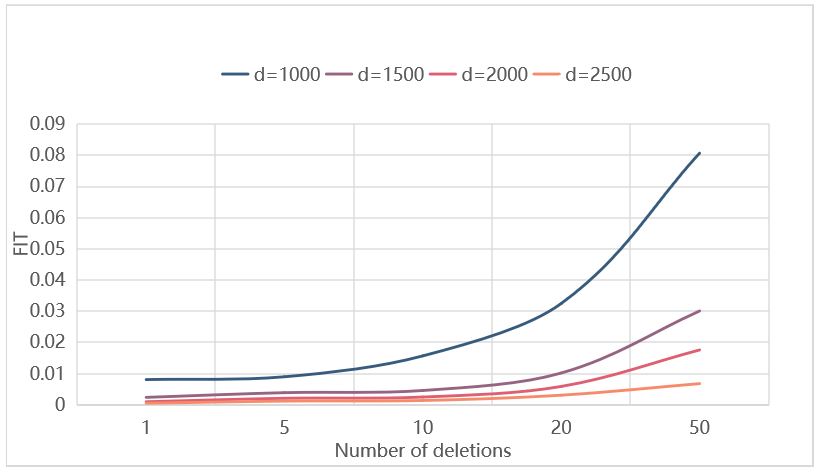}} 
\caption{\label{1} Discriminant model network framework} 
\end{figure}

In the introduction earlier in this article, we know that the evaluation of machine learning is mainly about efficiency and effectiveness. Combining the aforementioned experimental findings, we can observe that, in terms of time consumption, our technique performs much better than INF, SISA, and model retraining methods. Meanwhile, in terms of effectiveness, our method also outperforms INF and is similar to retraining. The effect of model retraining here is the desired effect of machine learning. The performance of INF is far from ideal in most cases and is much less stable numerically than the new method adopted in this paper.

Median baseline weights on injected feature can be obtained from Table 5.

\begin{table}[!htbp]
    \centering
    \caption{Mean results for the FIT}
    \begin{tabular}{cccc}
        \toprule
        k value & p = 0.5 	& p = 0.25 &  p = 0.1 \\
        \midrule
        k = 10	& 8.81	& 9.17	     & 10.84  \\
        k = 50 	& 11.56	& 10.33	     & 9.62  \\ 
        k = 100  &  9.39		& 	9.52	 &   10.03 \\
        \bottomrule
    \end{tabular}\label{tb:Fitness_formations5}
\end{table}

The experimental findings show that our method's parameter update time is much quicker than model retraining. especially when $d = 2500$ and a removal cluster of dimension $k = 1$ are present. It is anticipated that residual updates would be over a hundred times quicker than exact retraining. And according to the FIT results, our method approximates model retraining in the machine unlearning task. And outperforms the INF method.

The efficiency of our method is independent of the size of the dataset, and its execution time is linear in the data dimension. This is an advance in high-dimensional machine learning. At the same time, we extend it to scientific practice and use it for downstream specific tasks of natural language processing models.

\section{conclusion}

We propose the Newton iterative projected residual method to implement the machine unlearning requirements of linear regression and neural networks. The computational cost of this method is linear in the data dimension, which is much better than existing quadratic correlation methods. Experimental results show that the Newton iteration-based projected residual method is effective for high-dimensional data and few-object deletion problems. Without affecting the accuracy of the original model, the effect of this method is similar to that of the retraining method and is significantly less time-consuming. The number of target deletions has a major impact on how well this technique performs the machine learning job. For any approximate deletion approach, the accuracy of the estimate decays as the number of deleted requests processed grows. As a result, it is possible to confirm the viability of this approach.

\section{Acknowledgement}
This paper is supported by the Key Research and Development Program of Guangdong Province under grant No.2021B0101400003. Corresponding author is Jianzong Wang from Ping An Technology (Shenzhen) Co., Ltd (jzwang@188.com).

% \newpage

\bibliographystyle{IEEEtran}
\bibliography{mybib}

\vspace{12pt}

\end{document}